\pdfoutput=1
\documentclass[11pt]{article}
\usepackage{soul}
\usepackage{EMNLP2023}

\usepackage{times}
\usepackage{latexsym}
\usepackage{longtable}      % for multi-page tables
\usepackage{booktabs}       % for \toprule, \midrule, \bottomrule
\usepackage{tabularx}       % for X-column types

\usepackage{multirow}
\usepackage{array}

\usepackage{lipsum}
\usepackage[T1]{fontenc}
\usepackage[utf8]{inputenc}

\usepackage{times}
\usepackage{latexsym}

\usepackage{colortbl}

\usepackage{subfigure}
\usepackage{float}
\usepackage{graphbox}
\usepackage{svg}
\usepackage{nicefrac}
\usepackage{xcolor}

\usepackage{bold-extra}
\usepackage[T1]{fontenc}

\usepackage{array}
\newcolumntype{P}[1]{>{\centering\arraybackslash}p{#1}}
\newcolumntype{M}[1]{>{\centering\arraybackslash}m{#1}}

\definecolor{orange}{rgb}{1,0.5,0}
\definecolor{graynode}{RGB}{20,20,20}
\definecolor{crimsonred}{RGB}{220,20,60}
\definecolor{darkgraynode}{gray}{0.5}
\definecolor{lightgraynode}{gray}{0.8}

\usepackage{rotate}
\usepackage{adjustbox}
\usepackage{array}
\usepackage{capt-of}
\usepackage{tabulary}
\usepackage{setspace}
\usepackage{amssymb}
\usepackage{mathtools}
\usepackage{pifont}

\definecolor{gray}{RGB}{20,20,20}
\definecolor{gray}{RGB}{0.7,0.7,0.7}
\definecolor{greencm}{RGB}{0,153,0}

\definecolor{plotblue}{RGB}	{30,144,255}
\definecolor{plotgreen}{RGB}	{50,205,50}
\definecolor{plotred}{RGB}	{220,20,60}

\definecolor{myyellow}{RGB}{255,255,204}
\definecolor{myred}{RGB}{255,204,204}
\definecolor{myblue}{RGB}{0,200,255}
\definecolor{mygreen}{RGB}{80,220,80}

\newcommand*\hrulefillvar[1][0.4pt]{\leavevmode\leaders\hrule height#1\hfill\kern0pt}

\usepackage{algorithm}
\usepackage[noend]{algpseudocode}
\algrenewcommand\algorithmicrequire{\textbf{Input:}}
\algrenewcommand\algorithmicensure{\textbf{Output:}}

\usepackage{subfigure}
\usepackage{nicefrac}

\usepackage[normalem]{ulem} % Import ulem and keep italics normal

\DeclareMathAlphabet{\mathbcal}{OMS}{cmsy}{b}{n}
\usepackage{amsmath}
\usepackage{mathrsfs}
\usepackage{comment}

\usepackage{bm}
\usepackage{bbm}
\usepackage{amssymb}
\usepackage{tcolorbox}
\usepackage[fixed]{fontawesome5}

\usepackage{paralist}
\definecolor{thedarkblue}{RGB}{0,0,120} %104} % 180
\definecolor{mydarkblue}{rgb}{0,0.08,0.45} %ICML dark blue

\usepackage{hyperref}
\hypersetup{%
    colorlinks=true,
    linkcolor=mydarkblue,
    citecolor=mydarkblue,
    filecolor=mydarkblue,
    urlcolor=mydarkblue}

\definecolor{googleblue}{HTML}{4285F4}
\definecolor{googlered}{HTML}{DB4437}
\definecolor{googlepurple}{HTML}{A142F4} % New purple color
\definecolor{googlegreen}{HTML}{0F9D58}

\definecolor{googleyellow}{HTML}{F4B400} % Bright yellow color
\definecolor{googleorange}{HTML}{FBBC05} % Orange shade
\definecolor{googlecyan}{HTML}{34A853} % Light cyan-green
\definecolor{googlegray}{HTML}{9AA0A6} % Medium gray for text or subtle elements
\definecolor{googlepink}{HTML}{EA4335} % Reddish-pink hue
\definecolor{googlelightblue}{HTML}{7BAAF7} % Lighter blue for highlights or backgrounds

\usepackage{makecell}

\usepackage{inconsolata}
\NewDocumentCommand{\zihao}{ mO{} }{\textcolor{red}{\textsuperscript{\textit{zihao}}\textsf{\textbf{\small[#1]}}}}
\NewDocumentCommand{\Mo}{ mO{} }{\textcolor{teal}{\textsuperscript{\textit{Mo}}\textsf{\textbf{\small[#1]}}}}

\NewDocumentCommand{\ying}{ mO{} }{\textcolor{cyan}{\textsuperscript{\textit{ying}}\textsf{\textbf{\small[#1]}}}}

\NewDocumentCommand{\lifu}{ mO{} }{\textcolor{orange}{\textsuperscript{\textit{Lifu}}\textsf{\textbf{\small[#1]}}}}

\NewDocumentCommand{\samya}{ mO{} }{\textcolor{blue}{\textsuperscript{\textit{samya}}\textsf{\textbf{\small[#1]}}}}

\usepackage[nopatch=footnote]{microtype}
\title{A Survey on Mechanistic Interpretability for \\ Multi-Modal Foundation Models}

\author{Zihao Lin$^{1*}$ \quad Samyadeep Basu$^{2*}$ \quad  Mohammad Beigi$^1$\thanks{The first three authors are co-first authors with equal contributions: qzlin@ucdavis.edu, sbasu12@umd.edu, mbeigi@ucdavis.edu.} \\ \textbf{Varun Manjunatha}$^3$ \,
 \textbf{Ryan A. Rossi}$^3$ \,  \textbf{Zichao Wang}$^3$ \, \textbf{Yufan Zhou}$^3$ \\ \textbf{Sriram Balasubramanian}$^2$ \,
 \textbf{Arman Zarei}$^2$ \, \textbf{Keivan Rezaei}$^2$ \, \textbf{Ying Shen}$^4$ \, \textbf{Barry Menglong Yao}$^1$ \\ \textbf{Zhiyang Xu}$^5$  \, \textbf{Qin Liu}$^1$ \,
 \textbf{Yuxiang Zhang}$^6$  \, \textbf{Yan Sun}$^7$ \, \textbf{Shilong Liu}$^8$ \, \textbf{Li Shen}$^9$ \,  \textbf{Hongxuan Li}$^{10}$   \\  \textbf{Soheil Feizi}$^{2\dagger}$ \,  \textbf{Lifu Huang}$^1$\thanks{The last two authors are co-corresponding authors: sfeizi@cs.umd.edu, lfuhuang@ucdavis.edu. }\\
  $^1$UC Davis \, $^2$University of Maryland \, $^3$Adobe \, $^4$UIUC \, $^5$Virginia Tech \, $^6$Waseda University \\ \, $^7$University of Sydney \, $^8$Tsinghua University \, $^9$Sun Yat-Sen University \,  $^{10}$Duke University
  }

\begin{document}
\maketitle
\begin{abstract}
The rise of foundation models has transformed machine learning research, prompting efforts to uncover their inner workings and develop more efficient and reliable applications for better control. While significant progress has been made in interpreting Large Language Models (LLMs), multimodal foundation models (MMFMs)—such as contrastive vision-language models, generative vision-language models, and text-to-image models—pose unique interpretability challenges beyond unimodal frameworks. Despite initial studies, a substantial gap remains between the interpretability of LLMs and MMFMs. This survey explores two key aspects: (1) the adaptation of LLM interpretability methods to multimodal models and (2) understanding the mechanistic differences between unimodal language models and cross-modal systems. By systematically reviewing current MMFM analysis techniques, we propose a structured taxonomy of interpretability methods, compare insights across unimodal and multimodal architectures, and highlight critical research gaps.
\end{abstract}

\section{Introduction}

The rapid development and adoption of multimodal foundation models (MMFMs)—particularly those integrating image and text modalities—have enabled a wide range of real-world applications.
For example, text-to-image models ~\citep{rombach2022highresolutionimagesynthesislatent, ramesh2022hierarchicaltextconditionalimagegeneration, podell2023sdxlimprovinglatentdiffusion} facilitate image generation and editing, generative vision-language models (VLMs)~\citep{zhu2023minigpt4enhancingvisionlanguageunderstanding, agrawal2024pixtral12b} support tasks like visual question answering (VQA) or image captioning tasks, and contrastive (i.e., non-generative) VLMs such as CLIP~\citep{radford2021learningtransferablevisualmodels} are widely used for image retrieval.
As multimodal models advance, there is a growing need to understand their internal mechanisms and decision-making processes~\citep{basu2024understandinginformationstoragetransfer}. Mechanistic interpretability is crucial not only for explaining model behavior but also for enabling downstream applications such as model editing~\citep{basu2024understandinginformationstoragetransfer}, mitigating spurious correlations~\citep{balasubramanian2024decomposing}, and improving compositional generalization~\citep{zarei2024understanding}.

\textit{Interpretability} in machine learning, LLMs, and multimodal models is a broad and context-dependent concept, varying by task, objective, and stakeholder needs. In this survey, we adopt the definition proposed by \citet{murdoch2019definitions}: ``\textit{The process of extracting and elucidating the relevant knowledge, mechanisms, features, and relationships a model has learned, whether encoded in its parameters or emerging from input patterns, to explain how and why it produces outputs.}'' This definition emphasizes the extraction and understanding of model knowledge, but what constitutes relevant knowledge'' depends on the application context. For instance, in memory editing applications, interpretability enables precise modifications to internal representations without disrupting other model functions, while in security contexts, it helps highlight input features and activations that signal adversarial inputs. Through this lens, this survey examines interpretability methods, exploring how they uncover model mechanisms, facilitate practical applications, and reveal key research challenges.

While interpretability research has made significant progress in unimodal large language models (LLMs)~\citep{meng2022locating, marks2024sparsefeaturecircuitsdiscovering}, the study of MMFMs remains comparatively underexplored.
Given that most multimodal models are transformer-based, several key questions arise: {\it Can LLM interpretability methods be adapted to multimodal models}? If so, do they yield similar insights? {\it Do multimodal models exhibit fundamental mechanistic differences from unimodal language models}?
Additionally, to analyze multimodal-specific processes like cross-modal interactions, {\it are entirely new methods required}? Finally, we also examine the practical impact of interpretability by asking—{\it How can multimodal interpretability methods enhance downstream applications}?

To address these questions, we conduct a comprehensive survey and introduce a three-dimensional taxonomy for mechanistic interpretability in multimodal models: (1) \textbf{Model Family} – covering text-to-image diffusion models, generative VLMs, and non-generative VLMs; (2) \textbf{Interpretability Techniques} – distinguishing between methods adapted from unimodal LLM research and those originally designed for multimodal models; and (3) \textbf{Applications} – categorizing real-world tasks enhanced by mechanistic insights.

Our survey synthesizes existing research and uncovers the following insights: (i) LLM-based interpretability methods can be extended to MMFMs with moderate adjustments, particularly when treating visual and textual inputs similarly. (ii) Novel multimodal challenges arise such as interpreting visual embeddings in human-understandable terms, necessitating new dedicated analysis methods. (iii) While interpretability aids downstream tasks, applications like hallucination mitigation and model editing remain underdeveloped in multimodal models compared to language models. These findings can guide future research in multimodal mechanistic interpretability.

Recently, \citet{dang2024explainable} provides a broad overview of interpretability methods for MMFMs across data, model architecture, and training paradigms. Another concurrent work \cite{sun2024review} reviews the multimodal interpretability methods from a historical view, covering works from 2000 to 2025. While insightful, our work differs from theirs in both focus and scope. To be specific, our work examines how established LLM interpretability techniques adapt to various multimodal models, analyzing key differences between unimodal and multimodal systems in techniques, applications, and findings.

The \textbf{summary of our contributions} are:
\begin{itemize}
\item We offer a comprehensive survey of {\it mechanistic interpretability for multimodal foundation models} spanning generative VLMs, contrastive VLMs, and
%\lifu{update it accordingly}
text-to-image diffusion models.
\item We introduce a {\it simple and intuitive taxonomy} which helps to distinguish the mechanistic methods, findings, and applications across unimodal and multimodal foundation models, highlighting critical research gaps. % between unimodal and multimodal model interpretability.
\item Based on the mechanistic differences between LLMs and multimodal foundation models, we identify fundamental {\it open challenges and limitations} in multimodal interpretability, providing directions for future research
\end{itemize}

\begin{figure*}[ht!]
    \centering
    \includegraphics[width=1\textwidth]{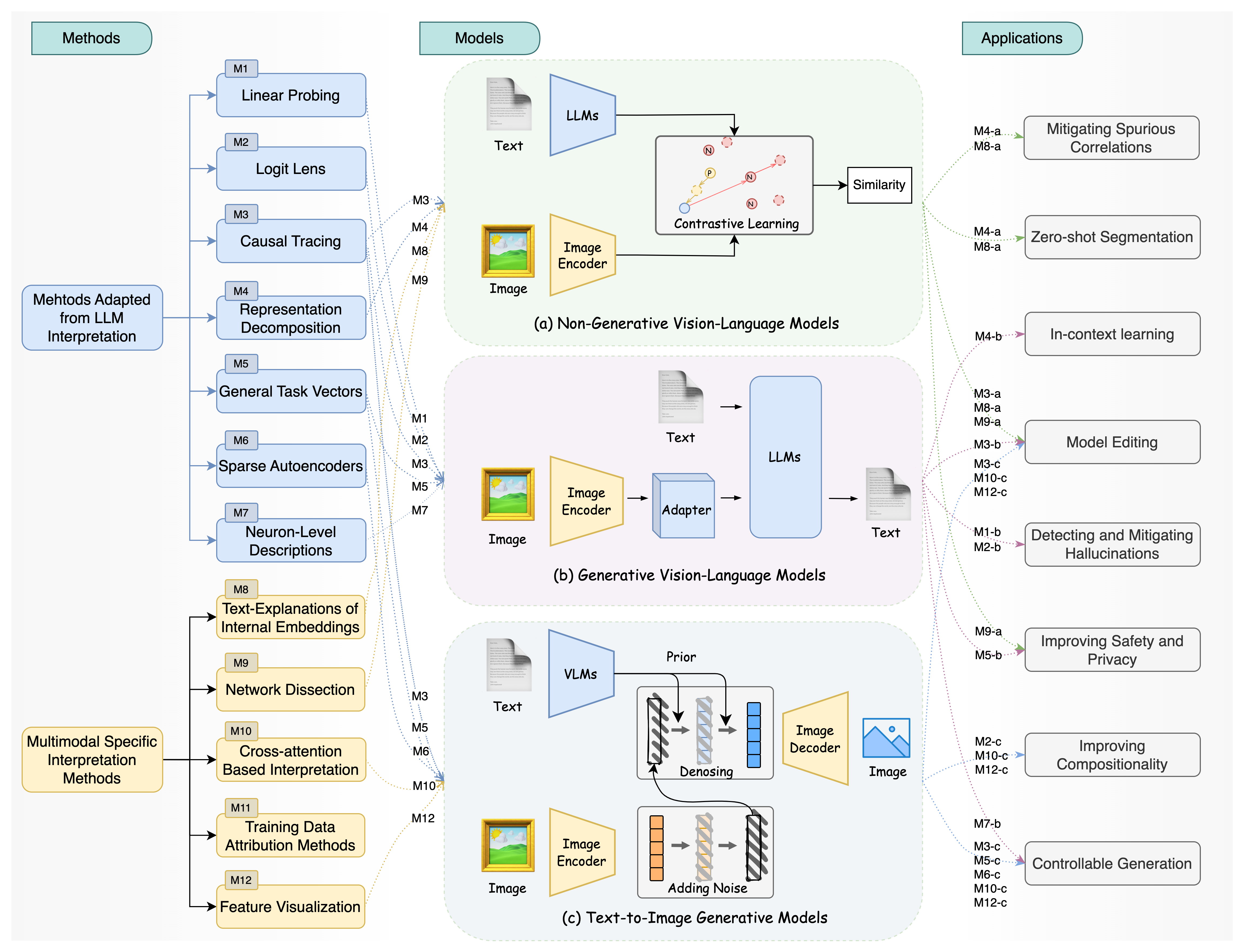}
    \caption{
    \textbf{In our survey, we study two types of mechanistic interpretability: (1) methods that adapted from LLM interpretability techniques and (2) multimodal-specific interpretability methods}. Different analysis methods are applied to three multimodal model architectures: (a) Non-generative Vision-Language Models, (b) Multimodal Large Language Models, and(c) Text-to-Image Generative Models (diffusion models especially). The interpretability insights from different methods and models can illuminate specific applications.
    }
    \vspace{-0.4cm}
    \label{fig:architecture}
\end{figure*}
\vspace{-0.5cm}

\section{Taxonomy}
In our survey, we present an easy-to-read taxonomy that categorizes mechanistic interpretability techniques along three dimensions: (i) Dimension 1 provides a view of the mechanistic insights across various multimodal model families including non-generative VLMs (e.g., CLIP), text-to-image models (e.g., Stable-Diffusion), and multimodal language models (e.g., LLaVa). We describe the architectures studied in our paper in Sec.(\ref{model_arch}); (ii) Dimension 2 categorizes whether the technique has been used for language models (Sec.\ref{llm_interpret}) or is specifically designed for multimodal models (Sec.\ref{non_llm_interpret});
(iii) Dimension 3 links insights from these mechanistic methods to downstream practical applications (Sec.\ref{applications}). The taxonomy is visualized in Figure \ref{fig:architecture}. In particular, the distribution of insights and applications are in-line in Sec. (\ref{llm_interpret}, \ref{non_llm_interpret}, \ref{applications}).

We believe this simple categorization will help readers (i) understand the gaps between unimodal language models and multimodal models in terms of mechanistic insights and applications, and (ii) identify the multimodal models where mechanistic interpretability (and their applications) is underexplored.

\section{Details on Model Architectures}
\label{model_arch}

In this section, we introduce three main categories of multimodal models covered by our survey, including (i) Contrastive (i.e., Non-Generative) Vision-Language Models, Generative Vision-Language Models, and Text-to-image Diffusion Models.  We choose these three families as they encompass the majority of the state-of-the-art architectures used by the community currently.

\subsection{Non-Generative Vision-Language Models}
\label{gvlm}
One non-generative vision-language model (e.g., CLIP~\citep{radford2021learningtransferablevisualmodels}, ALIGN~\citep{jia2021scalingvisualvisionlanguagerepresentation}, FILIP~\citep{yao2021filipfinegrainedinteractivelanguageimage}, SigCLIP~\citep{zhai2023sigmoidlosslanguageimage}, DeCLIP~\citep{li2022supervisionexistseverywheredata} and LLIP~\citep{lavoie2024modelingcaptiondiversitycontrastive}) usually contains one language-model-based text encoder and one vision-model-based vision encoder. These models are particularly suited for real-world applications such as text-guided image retrieval, image-guided text retrieval and zero-shot image classification.

\subsection{Text-to-Image Diffusion Models}
\label{t2i}
State-of-the-art text-guided image generation models are primarily based on the diffusion objective ~\citep{rombach2022highresolutionimagesynthesislatent, DBLP:journals/corr/abs-2006-11239}, which predicts the noise that was added during the forward diffusion process, allowing it to learn how to gradually denoise random Gaussian noise back into a clean image during the reverse diffusion process. One diffusion model often contains a text encoder (e.g., CLIP) and a CNN-based U-Net~\citep{journals/corr/RonnebergerFB15} for denoising to generate images. Early variants of text-to-image generative models with this objective include Stable-Diffusion-1~\citep{rombach2022highresolutionimagesynthesislatent} (which perform the diffusion process in a compressed latent space) and Dalle-2~\citep{ramesh2022hierarchicaltextconditionalimagegeneration} (which perform the diffusion process in the image space instead of a compressed latent space).
In recent times, SD-XL~\citep{podell2023sdxlimprovinglatentdiffusion} improves on the early Stable-Diffusion variants by using a larger denoising UNet and an improved conditioning (e.g., text or image) mechanism. More recent models such as Stable-Diffusion-3~\citep{esser2024scalingrectifiedflowtransformers} obtain stronger image generation results than previous Stable-Diffusion variants by (i) using a rectified flow formulation, (ii) scalable transformer architecture as the diffusion backbone and (iii) using an ensemble of strong text-encoders (e.g., T5~\cite{t5,flan-t5}). Beyond image generation, in terms of downstream applications, text-to-image models can also be applied for image editing~\citep{hertz2022prompt}, and style transfer~\citep{zhang2023inversionbasedstyletransferdiffusion}.

\subsection{Generative Vision-Language Models}
\label{mllm}
In our paper, we investigate the most common generative VLMs which are developed by connecting a vision encoder (e.g., CLIP) to a large language model through a bridge module. This bridge module (e.g., a few MLP layers \cite{liu2023visualinstructiontuning} or a Q-former \cite{li2023blip}) is then trained on large-scale image-text pairs. Frozen~\citep{tsimpoukelli2021multimodalfewshotlearningfrozen} is one of the first works to take advantage of a large language model in image understanding tasks (e.g., few-shot learning). Follow-up works such as MiniGpt~\citep{zhu2023minigpt4enhancingvisionlanguageunderstanding}, BLIP variants~\citep{li2023blip2bootstrappinglanguageimagepretraining} and LLava~\citep{liu2023visualinstructiontuning} improved on Frozen by modifying the scale and type of the training data, as well as the underlying architecture. In recent times, much focus has been geared toward curating high-quality image-text pairs encompassing various vision-language tasks. Qwen~\citep{yang2024qwen2technicalreport}, Pixtral~\citep{agrawal2024pixtral12b} and Molmo~\citep{deitke2024molmopixmoopenweights} are some of the recent multimodal language models focusing on high-quality image-text curated data. Multimodal language models have various real-world applications, such as VQA, and image captioning.

\textbf{Note}. We acknowledge the emergence of unified transformer-based multimodal models capable of both image generation and multimodal understanding, such as~\citep{xie2024show, team2024chameleon, dong2024dreamllmsynergisticmultimodalcomprehension}. However, we exclude these from our discussion due to the absence of mechanistic interpretability studies on them. Besides, another variant of model architecture, which is designed to generate interleaved images and text, such as GILL \cite{koh2024generating}, combines an MLLM and a diffusion model into one system. We will classify such a model based on its analyzed components.

\section{LLM Interpretability Methods for Multimodal Models}
\label{llm_interpret}
We first examine mechanistic interpretability methods originally developed for large language models and their adaptability to multimodal models with minimal to moderate modifications. Our focus is on \textit{\textbf{how existing LLM interpretability techniques can provide valuable mechanistic insights into multimodal models.}}

Specifically, we begin discussing diagnostic tools (Linear Probing (Sec. \ref{sec:linear-probing}), Logit Lens (Sec. \ref{sec:logit-lens})), which passively map what knowledge is encoded in model representations and where it resides across layers. We then introduce causal intervention methods (Causal Tracing and Circuit Analysis (Sec. \ref{sec:causal-tracing})), which actively perturb model states to uncover where the knowledge is stored and how specific predictions emerge in multimodal models. These insights then enable representation-centric approaches (Representation Decomposition (Sec. \ref{rep_decompose})) to mathematically disentangle activations into interpretable components, exposing the building blocks of model knowledge. This structural understanding directly informs behavioral control paradigms: General Task Vectors (Sec. \ref{sec:general-task-vectors}) leverage explicit task-driven arithmetic to edit model outputs, while Sparse Autoencoders (as their unsupervised counterpart, (Sec. \ref{sec:sparse-autoencoders})) provide machine-discovered feature bases for granular manipulation, bridging analysis to application. Finally, Neuron-level descriptions (Sec. \ref{sec:neuron-level-descriptions}) anchor these interpretations in empirical reality, validating macroscopic hypotheses through microscopic activation patterns (e.g., concept-specific neurons) and ensuring mechanistic fidelity.

\subsection{Linear Probing}
\label{sec:linear-probing}

\begin{figure*}[ht!]
    \centering
    \includegraphics[width=1\textwidth]{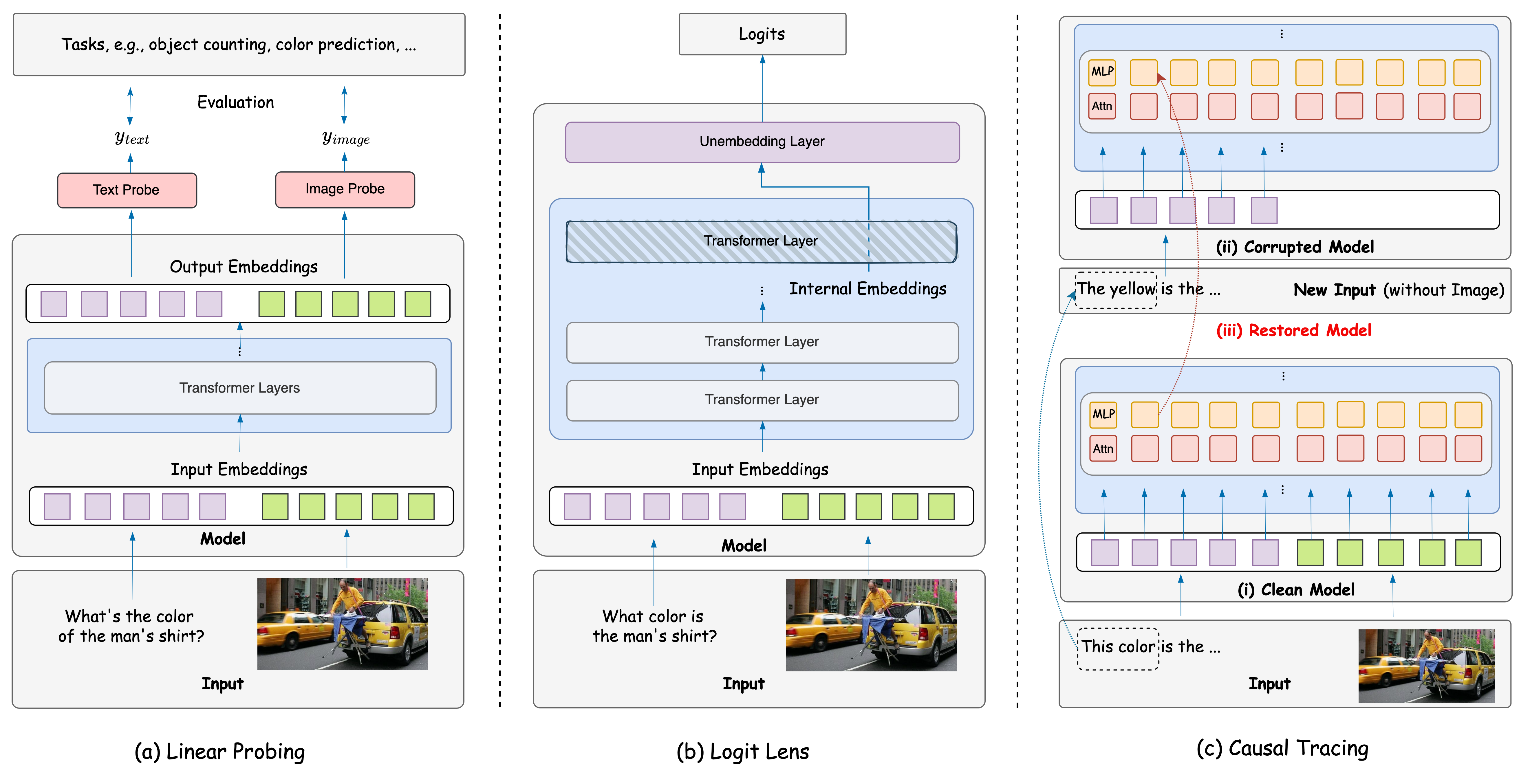}
    \caption{The illustrations of interpretability methods: (a) Linear Probing, (b) Logit Lens, and (c) Causal Tracing.}
    \vspace{-0.4cm}
    \label{fig:method-1-3}
\end{figure*}

Probing trains lightweight classifiers on {\it supervised}\footnote{The definition of supervision is described in Appendix \ref{app:more-definitions}.} probing datasets, typically linear probes, on frozen LLM representations to assess whether they encode linguistic properties such as syntax, semantics, and factual knowledge \cite{hao2021selfattentionattributioninterpretinginformation, liu2024probinglanguagemodelspretraining, zhang2024truthxalleviatinghallucinationsediting, liu2023cognitivedissonancelanguagemodel, beigi2024internalinspectori2robustconfidence}. The illustration of Linear Probing is shown in Figure \ref{fig:method-1-3} (a). This approach has been extended to multimodal models, introducing new challenges such as disentangling the relative contributions of each modality (i.e., visual or textual). To tackle these challenges, \citet{salin2022vision} developed probing methods to specifically assess how Vision-Language models synthesize and merge visual inputs with textual data to enhance comprehension, while \citet{dahlgren-lindstrom-etal-2020-probing} investigated the processing of linguistic features within image-caption pairings in visual-semantic embeddings.
Unlike in LLMs, where upper layers predominantly encode abstract semantics \cite{jawahar-etal-2019-bert, tenney-etal-2019-bert}, multimodal probing studies \cite{tao2024probingmultimodallargelanguage, salin2022vision} suggest that intermediate layers in multimodal models are more effective at capturing global cross-modal interactions, whereas upper layers often emphasize local details or textual biases. Furthermore, despite the fact that probing applications in LLMs are centered on specific linguistic analyses, the scope of probing in multimodal models extends to more varied aspects. For instance, \citet{dai2023plausiblefaithfulprobingobject} investigated object hallucination in vision-language models, analyzing how image encodings affect text generation accuracy and token alignment.

\begin{tcolorbox}
\vspace{-0.2cm}
\textbf{Main Findings and Gap.} The main drawback of linear probing is the requirement of supervised probing data and training a separate classifier for understanding concept encoding in layers. Therefore, scaling it via multimodal probing data curation and training separate classifiers across diverse multimodal models is a challenge.
\vspace{-0.2cm}
\end{tcolorbox}

\subsection{Logit Lens}
\label{sec:logit-lens}
The Logit Lens is an {\it supervised} interpretability method used to understand the inner workings of LLMs by examining the logits value of the output. As is shown in Figure \ref{fig:method-1-3} (b), this method conducts a layer-by-layer analysis, tracking logits at each layer (by projecting to the vocabulary space using the unembedding projection matrix) to observe how predictions evolve across the network. By decoding intermediate representations into a distribution over the output vocabulary, it reveals what the network ``thinks'' at each stage \cite{%nostalgebraist,
belrose2023eliciting}. In the context of multimodal models, studies show that predictions from earlier layers often exhibit greater robustness to misleading inputs compared to final layers~\cite{halawi2024overthinkingtruthunderstandinglanguage}. Studies also demonstrate that anomalous inputs alter prediction trajectories, making this method a useful tool for anomaly detection \cite{halawi2024overthinkingtruthunderstandinglanguage, belrose2023eliciting}. Additionally, for easy examples—situations where the model can confidently predict outcomes from initial layers—correct answers often emerge in early layers, enabling computational efficiency through adaptive early exiting \cite{schuster2022confidentadaptivelanguagemodeling, xin-etal-2020-deebert}.
Furthermore, the Logit Lens has been extended to analyze multiple inputs. \citet{huo2024mmneuron} adapted it to study neuron activations in feedforward network (FFN) layers, identifying neurons specialized for different domains to enhance model training. Further research has integrated contextual embeddings to improve hallucination detection \cite{phukan2024logitlenscontextualembeddings, zhao2024knowtokendistributionsreveal}. Additionally, the ``attention lens'' introduced in~\cite{jiang2024devilsmiddlelayerslarge} examines how visual information is processed, revealing that hallucinated tokens exhibit weaker attention patterns in critical layers.

\begin{tcolorbox}
\vspace{-0.2cm}
\textbf{Main Findings and Gap.} Beyond multimodal language models, logit-lens can be potentially utilised to mechanistically understand modern models such as unified understanding and generation models such as~\citep{xie2024show, team2024chameleon}.
\vspace{-0.2cm}
\end{tcolorbox}

\subsection{Causal Tracing}
\label{sec:causal-tracing}
Unlike passive diagnostic tools, Causal Tracing Analysis~\citep{pearl} is rooted in causal inference that studies the change in a response variable following an active intervention on intermediate variables of interest (mediators). An example of causal tracing applied to transformer-based generative VLM is illustrated in Figure \ref{fig:method-1-3} (c).
The approach has been widely applied to language models to pinpoint the network components—such as FFN layers—that are responsible for specific tasks \cite{meng2022locating, meng2022mass, 10.5555/2074022.2074073}.
For instance, \citet{meng2022locating} demonstrated that mid-layer MLPs in LLMs are crucial for factual recall, while \citet{stolfo2023mechanisticinterpretationarithmeticreasoning} identified the important layers for mathematical reasoning. Building on this technique and using a {\it supervised} probing dataset, \citet{basu2023localizing} found that, unlike LLMs, visual concepts (e.g., style, copyrighted objects) are distributed across layers in the noise model for diffusion models, but can be localized within the conditioning text-encoder. Further, \citet{basu2024mechanistic}
identified critical cross-attention layers that encode concepts like artistic style and general facts. Recent works have also extended causal tracing to mechanistically understand generative VLMs for VQA tasks~\citep{basu2024understandinginformationstoragetransfer, palit2023visionlanguagemechanisticinterpretabilitycausal, yu2024understandingmultimodalllmsmechanistic}, revealing key layers that guide model decisions in VQA tasks.
%We summarize the related papers in Table \ref{tab:causal-tracing}.
%\zihao{Should we summarize those papers in more sentences?}
\paragraph{Extending to Circuit Analysis} While causal tracing helps to identify individual ``causal'' components for a particular task, it does not automatically lead to the extraction of a sub-graph of the underlying computational graph of a model which is ``causal'' for a task. In this regard, there has been a range of works in language modeling to extract task-specific circuits~\citep{syed2023attributionpatchingoutperformsautomated, wang2022interpretabilitywildcircuitindirect, conmy2023automatedcircuitdiscoverymechanistic}. However, extending these methods to obtain task-specific circuits is still an open problem for MMFMs.

\begin{tcolorbox}
\vspace{-0.2cm}
\textbf{Main Findings and Gap.}
While causal tracing has been extensively used to analyze factuality and reasoning in LLMs, its application in multimodal models remains relatively limited. Expanding this method to newer, more complex multimodal architectures and diverse tasks remains an important challenge to address.
\vspace{-0.2cm}
\end{tcolorbox}

\subsection{Representation Decomposition}
\label{rep_decompose}
\begin{figure*}[ht!]
    \centering
    \includegraphics[width=1\textwidth]{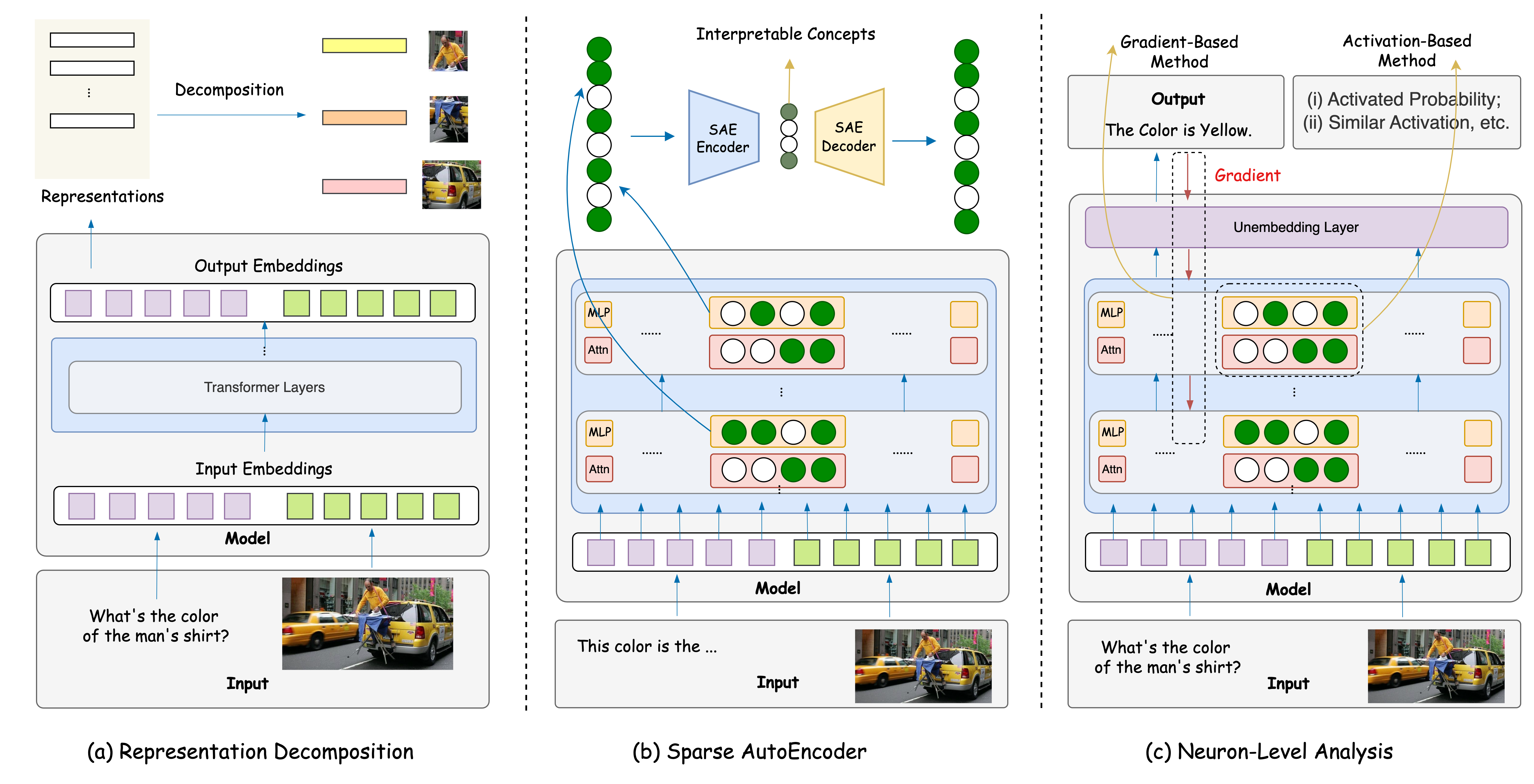}
    \caption{The illustrations of interpretability methods: (a) Representation Decomposition, (b) Sparse AutoEncoder, and (c) Neuron-level Analysis.}
    \vspace{-0.4cm}
    \label{fig:method-4-6}
\end{figure*}

In transformer-based LLMs, as illustrated in Figure \ref{fig:method-4-6}, the concept of representation decomposition pertains to the analysis of the model's internal mechanisms, specifically dissecting individual transformer layers to core meaningful components, which aims at understanding the inner process of transformers. In unimodal LLMs, research has mainly decomposed the architecture and representation of a model's layer into two principal components: the attention mechanism and the multi-layer perceptron (MLP) layer. Intensive research efforts have focused on analyzing these components to understand their individual contributions to the model's decision-making process. Studies find that while attention should not be directly equated with explanation \cite{pruthi2019learning, jain2019attention, wiegreffe2019attention}, it provides significant insights into the model's operational behavior and helps in error diagnosis and hypothesis development \cite{park2019sanvis, voita2019analyzing, vig2019visualizing, hoover-etal-2020-exbert, vashishth2019attention}. Furthermore, concurrently, research has shown that Feed-Forward Networks (FFNs) within the Transformer MLP layer, functioning as key-value memories, encode and retrieve factual and semantic knowledge ~\cite{geva2021transformer}. Experimental studies have established a direct correlation between modifications in FFN output distributions and subsequent token probabilities, suggesting that the model's output is crafted through cumulative updates from each layer~\cite{geva2022self}. This core property serves as the foundation for identifying language model circuits associated with specific tasks in~\citep{syed2023attributionpatchingoutperformsautomated, wang2022finding, conmy2023automated}.

In multimodal models, representation decomposition has been instrumental in analyzing modality processing and layer-specific properties. Studies such as ~\cite{gandelsman2024interpretingclipsimagerepresentation, balasubramanian2024decomposing} leverage {\it supervised} probing datasets and propose a hierarchical decomposition approach—spanning layers, attention heads, and tokens—to provide granular insights into model behavior.

Layer-wise decomposition reveals that shallow layers primarily integrate modality-specific inputs into a unified representation, while deeper layers refine task-specific details through denoising \cite{yin2024unraveling}. \citet{tao2024probingmultimodallargelanguage} further demonstrated that intermediate layers capture broader semantic information, balancing modality-specific details with holistic understanding—crucial for tasks such as visual-language entailment. In diffusion models like Stable Diffusion, \citet{prasad2023unravelingtemporaldynamicsunet} found that lower U-Net layers drive semantic shifts, while higher layers focus on denoising, progressively refining the latent representations into high-quality outputs.
\citet{quantmeyer2024doesclipprocessnegation} utilized causal tracing with representation decomposition to identify CLIP text encoder heads responsible for processing negation and semantic nuances, thereby improving cross-modal alignment.
Similarly, \citet{cao2020scenerevealingsecretspretrained} identified attention heads specialized for cross-modal interactions, integrating linguistic and visual cues for high-quality multimodal synthesis.
Notably, it shares similarities with causal tracing, which can be applied once a layer has been broken down into distinct components using Representation Decomposition.
\begin{tcolorbox}
\vspace{-0.2cm}
\textbf{Main Findings and Gap.} While CLIP and diffusion models are a great starting point for a case-study using representation decomposition, leveraging the inherent decomposability of transformers can be extended to understanding multimodal language models, and text-to-video models—an important gap that needs to be addressed.

\vspace{-0.2cm}
\end{tcolorbox}

\subsection{General Task Vectors}
\label{sec:general-task-vectors}
General Task (or steering) vectors in language models are directional embeddings that, when added to specific layers, enhance model capabilities such as in-context learning and instruction following.  To obtain these task vectors, one requires a well-annotated {\it supervised} probing dataset. \citet{hendel2023incontextlearningcreatestask} discovered a task vector for compressing task demonstrations, while \citet{zhang2024tellmodelattendposthoc} and \citet{jiang2024interpretable} leveraged instruction vectors to improve model adherence to user instructions and mitigate catastrophic forgetting.
In multimodal models, task vectors facilitate controlled image generation and editing. \citet{baumann2024continuoussubjectspecificattributecontrol} mapped text-embedding vectors to visual concepts for adjustable intensity, while \citet{gandikota2025concept} fine-tuned low-rank matrices in UNet to create controllable concept vectors. \citet{cohenmulti} explored multiple task vectors in diffusion models, proposing a prompt-conditioned adaptation method to minimize interference.

\begin{tcolorbox}
\vspace{-0.2cm}
\textbf{Main Findings and Gap.} While language models support both fine-tuning and zero-shot steering, multimodal models largely rely on fine-tuning. Advancing zero-shot steering for multimodal models remains a crucial research direction.
\vspace{-0.2cm}
\end{tcolorbox}

\subsection{Sparse Autoencoders: A Special Class of Unsupervised Task Vectors}
\label{sec:sparse-autoencoders}
Sparse Autoencoders (SAEs, \citet{yun-etal-2021-transformer}) offer an {\it unsupervised} approach to discovering conceptual representations in neural networks post-training. SAEs learn a dictionary of concepts such that any representation can be expressed as a linear combination of a \textit{sparse} subset of these concepts.
As illustrated in Figure \ref{fig:method-4-6} (b), an SAE is typically a two-layer MLP of the form  $SAE(x) = Dec( Act( Enc(x)))$ where $x$ is the input feature. The encoder ($Enc$) and the decoder ($Dec$) layers are simple linear layers and the activation function ($Act$) is a design choice and can be a simple ReLU~\citep{agarap2019deeplearningusingrectified}, Top K \cite{gao2024scalingevaluatingsparseautoencoders}, JumpReLU \cite{rajamanoharan2024jumpingaheadimprovingreconstruction}, and so on. The SAE is trained to reconstruct its own input, with the constraint that the activations should be sparse. Once trained, the neurons in the activation layer are assigned interpretations based on the highest activating input samples for the specific neuron in question. This results in a concept dictionary where concepts are mapped to directions (i.e., {\it vectors}) in representation space. These vectors can be added to the residual stream of the model to potentially control various facets such as the safety and intensity of various attributes in image generation models.
The SAE with an autoencoder architecture is trained to reconstruct its input while enforcing sparse activations. Once trained, neurons are interpreted based on their highest-activating inputs, forming a concept dictionary that maps concepts to vectors in representation space. These vectors can then be added to the model’s residual stream to control attributes like safety and intensity in image generation.
Due to their unsupervised nature, which minimizes the need for annotated examples for probing, SAEs have been applied extensively to LLMs to identify human interpretable directions for various concepts (e.g., refusal) in representation space \cite{cunningham2023sparseautoencodershighlyinterpretable}. These directions can then be used to steer the language model
\cite{marks2024sparsefeaturecircuitsdiscovering} without the need of fine-tuning it. More recently, SAEs have been extended to vision-language models like CLIP \cite{Daujotas2024, rao2024discover, lim2024sparseautoencodersrevealselective} and audio transcription models like Whisper \cite{Sadov2024}. Despite their promise, SAEs face challenges such as feature absorption and splitting \cite{chanin2024absorptionstudyingfeaturesplitting}, lack of robust evaluation metrics \cite{makelov2024principledevaluationssparseautoencoders} and underperformance compared to supervised methods for model control.

\subsection{Neuron-Level Descriptions}
\label{sec:neuron-level-descriptions}
Neuron-level analysis methods aim to identify specific neurons that contribute to model predictions~\citep{sajjad2022neuron}. The illustration is shown in Figure \ref{fig:method-4-6} (c). In this section, we divide these methods into two main categories: gradient-based attribution, and activation-based analysis.
%\footnote{Additional categories such as prediction probability changes and others are discussed in Appendix \ref{app:more-on-neuron-level-analysis}.}

There are different definitions of neurons in deep neural networks. We define $\mathbf{x}$ as the input embeddings, and $\mathbf{h_i}$ as the hidden states of the $i$-th layer's output. A model layer multiplies the hidden states with parameter $M_i$ followed by an activation function $\mathbf{a} = f(xM_i^\intercal)$. Some studies define the activation $a_j$, which is the $j$-th element of $\mathbf{a}$ as the neuron \cite{dai2021knowledge}. While other works \cite{dalvi2019one, durrani2020analyzing, antverg2021pitfalls} define the dimensions in output representation as a neuron. For consistency, in our survey, we follow the most widely used definition to define an element $m_j$ of a layer's parameter $M$ as the neuron.

%\zihao{@Samya, I moved all the methods from appendix to main body. Should we Keep them in appendix or it is fine to describe them here?}
\paragraph{Gradient-based attribution} methods analyze how neuron values influence model outputs by perturbing neuron activations and accumulating weight contributions based on corresponding gradients~\citep{dai2021knowledge}.
In unimodal settings, \citet{dai2021knowledge} detected fact-related neurons concentrated in the top layers of a pretrained language model, such as BERT~\citep{devlin2018bert},
while \citet{wang2022hint} identified neurons for encoding hierarchical concepts in a CNN-based vision model, such as VGG19~\citep{journals/corr/SimonyanZ14a}.
Extending this approach to multimodal settings, \citet{schwettmann2023multimodal} identified ``multimodal neurons'' that transform visual representations into textual concepts via the model’s residual stream.

\paragraph{Activation-based analysis} methods detect whether a neuron is activated when processing an input. These methods have been used to identify neurons specialized for specific tasks~\citep{wang2022finding} and multilingual understanding~\citep{tang2024language}. Additionally, \citet{voita2023neurons} identified "dead" neurons that are never activated, revealing the sparsity of LLMs. In multimodal contexts, \citet{goh2021multimodal} detected neurons encoding distinct visual features in non-generative models, while in generative VLMs, researchers have identified domain-specific neurons~\citep{huo2024mmneuron} and modality-specific neurons~\citep{huang2024miner}. In diffusion models, \citet{hintersdorf2024finding} identified memorization neurons by analyzing their out-of-distribution activations.

\paragraph{Prediction Probability Changes} methods usually change the neuron output value, and analyze its influence on the final prediction. \citet{yu2024neuron} quantifies the importance level of a neuron by calculating the difference of the log of the probabilities by giving and without giving the neuron value. In this way, this paper finds that both attention and FFN layer store knowledge. Besides, all important neurons directly contributing to knowledge prediction are in deep layers. \citet{yu2024interpreting} utilizes the same method to find that features are enhanced in shallow FFN layers and neurons in deep layers are used to enhance prediction. Following a similar strategy, \citet{yu2024understanding} finds important attention heads for handling VQA tasks.

\paragraph{Attribution Method} is to project the internal hidden representation into output space to analyze each neuron's contribution to the final prediction \cite{geva2022transformer}. In the multimodal domain, \citet{pan2023finding} projects the activation of one neuron into output space to quantify the importance of one neuron to the final prediction and identify multimodal neurons. \citet{fang2025towards} utilizes this method to find the semantic knowledge neurons and some interesting properties such as cross-modal invariance and semantic sensitivity.

\paragraph{Other Method} covers many different types of neuron-level analysis methods. For example, instead of directly analyzing the first-order effect, which is the logits of each neuron, \citet{gandelsman2024interpreting} analyzes the accumulation of information of a neuron after the attention head. A new method to analyze information flow. Focus on the contribution of neurons to the output representation.

\begin{tcolorbox}
\vspace{-0.2cm}
\textbf{Main Findings and Gap.}
Neuron-level analysis adapts well to multimodal settings, but deeper neuron interactions remain underexplored, such as activation shifts in generative VLMs when adding visual input to identical text.
\vspace{-0.2cm}
\end{tcolorbox}

\subsection{Summary}
Overall, we find that the core principles of popular LLM-based mechanistic interpretability methods can be extended to multimodal models without complex modification. However, extracting meaningful mechanistic insights from these models often requires carefully tailored adaptations.

\begin{tcolorbox}
\vspace{-0.2cm}
\textbf{Main Findings and Gap.} The effectiveness of SAEs as a control mechanism for multimodal models is still in its early stages and requires validation across a range of multimodal models, including the latest diffusion models and MLLMs.
\vspace{-0.2cm}
\end{tcolorbox}

\section{Interpretability Methods Specific to Multimodal Models}
\label{non_llm_interpret}

Many recent research studies also propose multimodal-specific inner mechanism interpretation analysis methods. Different from LLM-based methods introduced in Sec. \ref{llm_interpret}, those methods are designed and applied only for multimodal foundation models. These methods include techniques for annotating embeddings or neurons in human-understandable language (Sec. \ref{sec:text-explanations-of-internal-embeddings} and Sec. \ref{sec:network-dissection}, leveraging unique multimodal architectural components like cross-attention layers for deeper insights (Sec. \ref{app:cross-attn}), developing novel data attribution methods tailored to multimodal models, such as text-to-image diffusion models (Sec. \ref{sec:training-data-attribution-methods}) and specific visualization methods (Sec. \ref{sec:feature-visualization}).

\subsection{Text-Explanations of Internal Embeddings}
\label{sec:text-explanations-of-internal-embeddings}
In Sec. \ref{rep_decompose}, we leverage the representation decomposition property of transformers to identify key components in token representations. However, interpreting these components in human-understandable terms remains a challenge.
For CLIP models, \citet{gandelsman2024interpretingclipsimagerepresentation} proposed TextSpan, which assigns textual descriptions to model components (e.g., attention heads) by identifying a text embedding that explains most of the variance in their outputs. The dataset for this task is {\it supervised} in nature. Expanding on this, \citet{balasubramanian2024decomposing} introduced a scoring function to rank relevant textual descriptions across components. Concurrently, SpLiCE \citep{bhalla2024interpreting} mapped CLIP visual embeddings to sparse, interpretable concept combinations. Additionally, \citet{parekh2024concept} employed dictionary learning to show that predefined concepts are semantically grounded in both vision and language.
Together, these methods enhance the interpretability of internal embeddings in multimodal models by providing textual explanations.
All the text-explanations of internal embedding papers aim to interpret where knowledge is stored in the model.

\begin{tcolorbox}
\vspace{-0.2cm}
\textbf{Main Findings and Gap.}
Current text-based explanations of internal embeddings primarily focus on simple concepts (e.g., color, location). It remains unclear whether these methods can effectively map visual embeddings to more abstract concepts, such as physical laws. Moreover, their applicability beyond CLIP, particularly in text-to-image and video generation models, remains largely underexplored.
\vspace{-0.2cm}
\end{tcolorbox}

\subsection{Network Dissection}
\label{sec:network-dissection}
Network Dissection (ND)~\citep{bau2017network}, pioneered automated neuron interpretability in multimodal networks by establishing connections between individual neurons and human-understandable concepts. Different from the internal embedding methods (Sec. \ref{sec:text-explanations-of-internal-embeddings}), ND compares %this method works by comparing
neuron activations with groud-truth concept annotations in images. When a neuron's activation pattern consistently matches with a specific concept over a certain threshold, that concept is assigned as the neuron's interpretation \cite{oikarinen2023clipdissect, falcon}. Moving beyond simple concept matching, MILAN \cite{hernandez2021natural} introduced a generative approach that produces natural language descriptions of neuron behavior based on highly activating images.
DnD~\citep{bai2024describe} then extend this work by first leveraging a generative VLM to describe highly activating images for each neuron and semantically combine these descriptions using an LLM.

\vspace{-0.2cm}
\begin{tcolorbox}
\textbf{Main Findings and Gap.}
The generalization of this method are constrained by their underlying multimodal architectures, e.g., CLIP. Moreover, while ND has proven effective for CNN-based vision models, its applicability to more advanced architectures, e.g., diffusion models, remains unexplored.
\vspace{-0.2cm}
\end{tcolorbox}

\subsection{Cross-attention Based Interpretability}
\label{app:cross-attn}
Cross-attention layers are crucial in multimodal models such as text-to-image diffusion models and generative VLMs, as they mediate interactions between image and text modalities. In generative models, studies have shown that cross-attention layers in UNet or DiT backbones play a critical role in linking an image's spatial layout to each word in the prompt \citep{tang2022daam}.
Building on this, \citet{hertz2022prompt} introduced a method for image editing via cross-attention control, enabling localized modifications, attribute amplification, and global changes while preserving image integrity. Similarly, \citet{neo2024interpretingvisualinformationprocessing} identified memorization neurons within cross-attention layers, while \citet{basu2024localizing} found that key concepts—such as artistic style, and factual knowledge—are concentrated in a small subset of these layers.

\begin{tcolorbox}
\vspace{-0.2cm}
\textbf{Main Findings and Gap.} While the cross-attention mechanisms in U-Net-based diffusion models are well-studied for applications like image editing and compositionality, mechanistic analysis of cross-attention in diffusion transformers (DiTs) and generative VLMs for downstream applications remains an open research area.
\vspace{-0.2cm}
\end{tcolorbox}

\subsection{Training Data Attribution Methods}
\label{sec:training-data-attribution-methods}
Training data attribution identifies training examples crucial to a specific prediction or generation. Although well studied for non-generative vision models \citep{koh2020understandingblackboxpredictionsinfluence, basu2021influencefunctionsdeeplearning, DBLP:conf/nips/PruthiLKS20@data-attribution-dm3, DBLP:conf/icml/ParkGILM23@data-attribution-gm4}, extending these methods to generative multimodal models (e.g., diffusion, multimodal language) remains challenging. Here, we highlight three categories of approaches specific to text-to-image diffusion models, with some other categories in the last paragraph.
%\ziaho{@Samya, need to revise this sentence.}

\paragraph{Retrieval and Unlearning Based Methods} A major challenge in training data attribution for diffusion models is the costly retraining needed for ground-truth influence and the adaptation of attribution methods due to time-step dependence. \citet{wang2023evaluating} evaluated retrieval-based attribution using image encoders (e.g., CLIP) as a baseline but did not incorporate diffusion model parameters. To address this, \citet{wang2024dataattributiontexttoimagemodels} introduced an unlearning-based approach, where generated images are ``unlearned'' by increasing their loss, creating an unlearned model. Attribution is then measured based on the deviation in training loss between the original and unlearned models, showing a strong correlation with ground-truth attribution.

\paragraph{Gradient-Based Methods} which are vital for data attribution in multimodal models, quantifying how training samples influence outputs via gradients.
For diffusion models, adaptations include K-FAC \cite{mlodozeniec2024influencefunctionsscalabledata@data-attribution-gm3}, which approximated the Generalized Gauss-Newton (GGN) matrix for scalable influence estimation, TRAK \cite{DBLP:conf/icml/ParkGILM23@data-attribution-gm4}, which modeled networks as kernel machines for improved attribution accuracy, and D-TRAK \cite{DBLP:conf/iclr/ZhengPD0L24@data-attribution-gm5}, which leveraged reverse diffusion and optimized gradient features for enhanced robustness. Additionally, DataInf \cite{DBLP:conf/iclr/KwonWW024@data-attribution-pm7} bridged perturbation methods with influence function approximations. Collectively, these techniques refine gradient-based attribution by disentangling multimodal attribution patterns through targeted perturbations.

\paragraph{Training Dynamics-Based Methods}
These methods analyze how model parameters and predictions evolve during training to determine the influence of specific data points, thereby revealing how models learn from and prioritize instances.
However, applying them to multimodal or generative models—like diffusion models—poses challenges.
For instance, Training Data Influence (TracIn)~\cite{DBLP:conf/nips/PruthiLKS20@data-attribution-dm3} can suffer from “timestep-induced bias,” where varying gradient magnitudes exaggerate the influence of some samples.
Diffusion-ReTrac~\cite{DBLP:journals/tmlr/XieLBH24@data-attribution-dm1} mitigates this by normalizing influence contributions.
Additionally, methods not originally designed for data attribution, such as CLAP4CLIP~\cite{jha2024clap4clipcontinuallearningprobabilistic@data-attribution-dm2} for VLMs, can still provide valuable insights through components like memory consolidation, weight initialization, and task-specific adapters that highlight crucial data points during training.

\paragraph{Other Miscellaneous Methods}
By contrasting similar and dissimilar data, these techniques trace how training examples influence model outputs.
For example, one approach fine-tunes a pre-trained text-to-image model using exemplar pairs and employs NT-Xent loss to generate soft influence scores~\cite{DBLP:conf/iccv/WangEZ0234@data-attribution-clm1}.
Similarly, Data Adaptive Traceback (DAT)~\cite{DBLP:conf/aaai/PengZYZQ24@data-attribution-clm2} aligns pre-training examples with downstream performance in a shared embedding space.
Moreover, adversarial attack studies~\cite{wang2024exploringtransferabilitymultimodaladversarial@data-attribution-clm3} demonstrate that intra-modal contrastive learning can be used to distinguish between adversarial and benign samples, while cross-modal loss highlights features critical for image-text alignment.

\begin{tcolorbox}
 \vspace{-0.2cm}
\textbf{Main Findings and Gap.}
Multimodal data attribution is challenging due to the scale of heterogeneous pre-training data and complex model architectures, making retraining infeasible and inference slow. Efficient attribution methods and retraining-free evaluation techniques remain an open problem.
\vspace{-0.2cm}
\end{tcolorbox}

\subsection{Feature Visualizations}
\label{sec:feature-visualization}
In MMFMs, feature visualization techniques
typically involve generating heatmaps of gradients or relevance scores over input images, providing an intuitive way to understand which features contribute to a model’s final prediction.

\paragraph{Visualizing Relevance Scores} For a given prediction, \citet{robnik2008explaining} visualizes a relevance score of each feature by examining how the prediction changes if the feature is excluded, calculated as the probability difference before and after excluding the feature. \citet{zintgraf2017visualizing} enhances this model by considering spatial dependence, proposing that a pixel's impact is strongly influenced by its neighboring pixels, thus expanding from pixel-level to patch-level relevance and measuring feature influences from hidden layers. \citet{chefer2021transformer} further improves the method of accumulating relevance across multiple layers by introducing a relevance propagation rule. Another line of work involves training a separate explanation model to predict feature relevance scores and then visualize them. \citet{ribeiro2016should} train an explanation model to evaluate the contribution of each image patch or word to the prediction. \citet{park2018multimodal} collect two new datasets to train a multimodal model that can jointly generate visual attention masks to localize salient regions and region-grounded text rationales. \citet{lyu2022dime} extends the work of \cite{ribeiro2016should} by developing a more detailed analysis framework. They decompose a multimodal model into unimodal contributions (UC) and multimodal interactions (MI), and then apply \cite{ribeiro2016should} method to learn relevance scores for each feature based on these unimodal contributions and multimodal interactions. \citet{liang2022multiviz} further extends to be a four-stage interpretation framework: unimodal importance, cross-modal interactions, multimodal representations, and multimodal prediction.

\paragraph{Visualizing Gradient} Grad-CAM~\cite{selvaraju2017grad} firstly visualized a coarse localization map by tracking how gradients from a target concept (such as 'dog' in classification or word sequences in captioning) flow back to the final prediction layer, highlighting key mage regions responsible for concept prediction. For both non-generative VLMs and MMFMs, this method has been employed to visualize grounding capabilities~\citep{rajabi2024q} and information flow in multimodal complex reasoning tasks~\citep{zhang2024redundancy}.
For diffusion models, \citet{tang2022daam} aggregated cross-attention word–pixel scores within the denoising network to compute global attribution scores, thus showing how specific words in a text prompt influence different parts of a generated image. Instead of visualizing only the final generated images, \citet{park2024explaining} provided a more detailed view by visualizing regions of focus and the attention given to concepts from prompts at each denoising step.

\begin{tcolorbox}
\vspace{-0.2cm}
\textbf{Main Findings and Gap.}
While feature visualization methods have been successfully applied to simple tasks such as image classification and visual question answering (VQA), their adaptation to more complex tasks—such as long-form image-to-text generation—remains underexplored.
\vspace{-0.2cm}
\end{tcolorbox}

\subsection{Summary}
In this section, we explore methods designed specifically to analyze the inner workings of multimodal models. Our findings reveal that the internal embeddings and neurons of models like CLIP can be interpreted using human-understandable concepts. Additionally, the cross-attention layers in text-to-image diffusion models provide valuable insights into image composition. For training data attribution and feature visualization, we observe that existing techniques for vision models have been effectively adapted for multimodal models.

\section{Applications using Mechanistic Insights for MMFMs}
\label{applications}

In this section, we emphasize the downstream applications inspired by interpretability analysis methods described in Sec. (\ref{llm_interpret}) and Sec. (\ref{non_llm_interpret}). We first introduce in-context learning in Sec. \ref{sec:app-in-context-learning}, followed by model editing (Sec. \ref{app:model_edit}) and hallucination detection (Sec. \ref{app:hallucination}). Then we summarize the applications for improving safety and privacy in MMFMs in Sec. \ref{safety_privacy} and improving compositionally in Sec. \ref{app:compositionality}. Finally, we also list several other types of applications in Sec. \ref{other}.

\subsection{In-context Learning}
\label{sec:app-in-context-learning}
Introduced in Sec. \ref{sec:general-task-vectors}, \citet{hendel2023context} and \citet{liu2023context} establish that ICL in language models can be viewed through the lens of task vectors. Following these works, \citet{huang2024multimodal} characterizes multimodal task vectors as pairs of attention head activations and indices and applies those task vectors to generative VLMs in in-context learning settings to compress long prompts that would otherwise not fit in limited context length. \citet{luo2024task} further analyzes the transferability of task vectors from different modalities, which extends the application of task vectors.

\begin{tcolorbox}
\vspace{-0.2cm}
\textbf{Main Findings and Gap.} Can task vectors be applied to more complex in-context learning tasks still remains unexplored.
\vspace{-0.2cm}
\end{tcolorbox}

\subsection{Model Editing}
\label{app:model_edit}
\paragraph{Editing Localized Layers in Diffusion Models.}
Building on \citet{orgad2023editing}, which edits cross-attention layers by modifying key and value matrices, \citet{basu2024mechanistic} localize cross-attention layers responsible for specific visual attributes and propose a targeted editing method. They identify critical layers using a brute-force approach, intervening in a subset of cross-attention inputs and measuring effects on generation. Significant changes in the visual attribute highlight the relevant layers. Their method demonstrates that knowledge of artistic styles, facts, and trademark objects is concentrated in a few cross-attention layers, enabling efficient, scalable, and generalizable edits across text-to-image models. \citet{basu2023localizing} extends the causal mediation analysis from \cite{meng2022locating} to text-to-image models, identifying key layers in the U-Net and text encoder responsible for generating specific visual attributes. Unlike large language models, where causal layers are typically mid MLP layers and vary by knowledge type, they find that in text-to-image models, the first self-attention layer of the text encoder is the sole causal state. This insight enables an efficient model editing method by focusing modifications on this critical layer.
\paragraph{Editing MLLMs.}
\citet{basu2024understandinginformationstoragetransfer} employs causal tracing to identify key causal layers in multimodal language models like Llava~\citep{liu2023visualinstructiontuning} for a factual VQA task. These layers are then modified using a closed-form solution to incorporate long-tailed information or correct erroneous responses. While \citet{pan2023finding} benchmarks model editing methods from the language model literature, we note that these techniques do not leverage mechanistic insights.

\begin{tcolorbox}
\vspace{-0.2cm}
\textbf{Main Findings and Gap.} When compared to language models, large-batch and sequential model editing~\citep{lin2024navigating} are two underexplored areas in multimodal model editing.
\vspace{-0.2cm}
\end{tcolorbox}

\subsection{Detecting and Mitigating Hallucinations}
\label{app:hallucination}
\citet{dai2023plausiblefaithfulprobingobject} examines how image encodings (e.g., region, patch, grid) and loss functions impact hallucinations in contrastive and generative VLMs, proposing a lightweight fine-tuning method to mitigate them. \citet{jiang2024interpretingeditingvisionlanguagerepresentations} finds that hallucinated objects have lower confidence when projected onto the output vocabulary, using this insight to develop a feature editing algorithm that removes them from captions. \citet{jiang2024devilsmiddlelayerslarge} shows that real object tokens receive higher attention weights from visual tokens than hallucinated ones. \citet{cohen2024performancegapentityknowledge} further analyzes visual-to-text information flow, offering insights for hallucination detection. \citet{phukan2024logitlenscontextualembeddings} identifies logit lens limitations and introduces a similarity metric based on middle-layer embeddings to detect hallucinations. Overall, hallucination detection in MMFMs remains less explored compared to language models~\citep{sakketou-etal-2022-factoid, li2024dawndarkempiricalstudy, chen2024factchdbenchmarkingfactconflictinghallucination, cheng2023evaluatinghallucinationschineselarge, li2023haluevallargescalehallucinationevaluation, manakul2023selfcheckgptzeroresourceblackboxhallucination}.
We also find that there is a lack of reliable benchmarking for hallucination detection methods for multimodal language models, when compared to language models.

\begin{tcolorbox}
\vspace{-0.2cm}
\textbf{Main Findings and Gap.} There is a lack of reliable benchmarking and evaluation for hallucination detection methods for multimodal language models, when compared to language models.
\vspace{-0.2cm}
\end{tcolorbox}

\subsection{Improving Safety and Privacy}
\label{safety_privacy}

\subsubsection{Safety}
Early efforts to improve generative VLMs safety relied on fine-tuning~\citep{zong2024safetyfinetuningalmostcost}, but recent work leverages mechanistic tools (Sec. \ref{llm_interpret}, \ref{non_llm_interpret}). Task vectors enhance safety by ablating harmful directions during inference~\citep{wang2024steeringawayharmadaptive}, while SAEs enforce sparsity to disentangle harmful features \cite{sharkey2022taking, templeton2024scaling}. \citet{xu2025crossmodal} identifies hidden states crucial to safety mechanisms but find misalignment between modalities, proposing localized training to address it. In text-to-image models, SAEs help remove unwanted concepts \cite{cywiński2025saeuroninterpretableconceptunlearning, ijishakin2024hspace}, and interpretable latent directions improve safe generations \cite{li2024selfdiscoveringinterpretablediffusionlatent}. For non-generative VLMs like CLIP, most work fine-tunes models for safety~\citep{poppi2024safeclipremovingnsfwconcepts}, though interventional methods in~\citep{basu2023localizing, gandelsman2024interpretingclipsimagerepresentation} could help identify safety-related layers.

\subsubsection{Privacy}
\paragraph{Data Leakage through Attacks on Specific Modalities} Multimodal data privacy refers to the protection of privacy when handling data from multiple modalities, such as text, images, audio, and video. Since multimodal models process information from different sources, each modality may involve different types of sensitive data, making privacy protection more complex and crucial \cite{zhao2024survey}. Traditional data privacy aims to protect original data from leakage by isolating and encrypting it through restricted secure access, especially for the large foundation models \cite{rao2023building}. Therefore, technologies such as federated learning \cite{li2020federated} and differential privacy \cite{dwork2006differential} can still work well for general training. However, due to the tight interconnections between multimodal data, this means that a reverse attack using data from a specific modality could still lead to the leakage of data from other modalities, which has become a major challenge in multimodal data privacy. \citet{ko2023practical} focuses on similar issues, where data leakage can occur through membership attacks. In this paper, we further summarize the privacy attributes of multimodal data and define it as cross-modal privacy. Caused by the asymmetry of the knowledge contained in multimodal data, if attackers steal data from certain key modalities, it may be sufficient to reconstruct all the information, ultimately leading to data leakage. Recent work has focused on multimodal information measurement techniques \cite{zhao2024survey,liu2024safety}, which enhance privacy protection by quantifying the correlations between data from different modalities. It significantly strengthens local privacy and effectively reduces the leverage risk in MMFMs.

\paragraph{Privacy Leakage through Cross-modal Access} Direct data leakage is typically catastrophic, but such cases are rare in practical scenarios. A more common challenge of privacy leakage occurs during the training process \cite{fang2024privacy}. Reverse attacks on models for specific modalities can also lead to data leakage. \citet{liu2024survey} explore the risk in vision-language models and highlight the risks that reverse attacks on multi-modal aggregation can potentially lead to the recovery of image data. The same, this type of attack can also be initiated by the trainer, who may construct partially falsified training data to reverse-query the corresponding data from other modalities \cite{xu2024fakeshield}. To prevent such privacy leakage, a key technique is feature perturbation. By adding lightweight noise, it ensures that during multimodal information fusion, knowledge from cross-modal data cannot be easily mapped independently. This enhances the privacy level in the training process.

\paragraph{Unreliable Samples: Poisoning Attacks} Poisoning attacks pose a significant threat to data reliability, targeting the training process by injecting maliciously altered data into the system. These attacks manipulate the training data to introduce vulnerabilities, potentially causing models to produce inaccurate predictions or exhibit unintended behaviors. Attackers usually craft subtle changes but significantly impact model performance. In multimodal models, apart from the traditional poisoning of tampering with the original data, altering the mapping relationships has become another critical attack vector. \citet{liu2024multimodal} learn the impact of asymmetric data attacks on model training is significant, as even a small amount of manipulated data can cause a severe decline in model performance. This also leads to more severe backdoor attacks, where attackers can execute the attack without the need for additional information injection \cite{liu2024compromising,yang2024efficient}. Aimed to these attacks, an effective solution is to generate adversarial examples for evaluation. By evaluating the symmetry of the modalities and the mapping relationships, toxic samples can be avoided from harming the network during training.

\subsection{Improving Compositionality}
\label{app:compositionality}
Compositionality in text-to-image models refers to their ability to correctly represent object compositions, attributes, and relationships from a given prompt. \citet{huang2023t2icompbench} introduces a benchmark to assess compositionality challenges in these models. LayoutGPT \cite{feng2024layoutgpt} leverages LLMs with few-shot learning to generate bounding boxes, guiding diffusion models via pixel-space loss. Grounded Compositional Generation \cite{phung2024grounded} refines this by defining the loss in cross-attention space, improving performance. Similarly, \citet{rassin2024linguistic} enhances attribute correspondence by aligning object-attention maps with adjectives.
Beyond diffusion model modifications, some works address compositionality issues by improving text conditioning. \citet{zarei2024understanding} identifies erroneous attention in CLIP, where nouns misalign with adjectives, and proposes a projection layer to enhance attribute binding. Likewise, \citet{zhuang2024magnet} introduces a zero-shot method that adjusts object embeddings to strengthen relevant attribute associations while minimizing irrelevant ones.

\begin{tcolorbox}
\vspace{-0.2cm}
\textbf{Main Findings and Gap.} While compositionality is well-studied in diffusion models, extending this analysis to newer models like Flux remains an open research direction.\vspace{-0.2cm}
\end{tcolorbox}

\subsection{Other Relevant Applications}
\label{other}
In this section, we highlight some of the other relevant applications using mechanistic insights for multimodal models:

\paragraph{Controlled Image Generation and Editing} In text-to-image diffusion models, task vectors can be used to control and edit the intensity of a specific concept in an image~\citep{baumann2024continuoussubjectspecificattributecontrol, gandikota2025concept}, while keeping other parts of the image unchanged. For example, given the prompt ``{\it An image of a boy in front of a cafe}'', if the size of the boy's eyes needs to be increased, a task vector corresponding to eye size is added to the model to modify the visual concept of the eyes. In the case of image editing, ~\citep{hertz2022prompt} intervenes on the interpretable cross-attention features to incorporate text-guided image edits.

\paragraph{Zero-shot Segmentation and Mitigating Spurious Correlations} The Representation Decomposition framework~\citep{gandelsman2024interpretingclipsimagerepresentation, balasubramanian2024decomposing} enables mapping the contributions of different visual tokens to the final [CLS] token. This decomposed information can be ranked based on CLIP similarity to identify the most important tokens for a specific visual concept. These selected tokens then form the segment representing the given concept.
This framework when combined with Text-Explanations of Internal Components (see Sec.\ref{sec:text-explanations-of-internal-embeddings}), can also mitigate spurious correlations. For e.g., certain attention heads can be identified that encode spurious attributes (e.g., water when classifying waterbirds). By ablating the contributions of these attention heads to the final [CLS] token in the image encoder, spurious correlations in CLIP models can be partially mitigated.
\label{app:controlled}

\section{Tools and Benchmarks}
\label{ref:tools-and-benchmarks}
There are many interpretability tools for LLMs covering attention analysis \cite{nanda2022transformerlens, fiottokaufman2024nnsightndifdemocratizingaccess}, SEA analysis \cite{bloom2024saetrainingcodebase}, circuit discovering \cite{, conmy2023automated}, causal tracing \cite{wu2024pyvene}, vector control \cite{vogel2024repeng, zou2023transparency}, logit lens \cite{belrose2023eliciting}, and token importance \cite{NIPS2017_7062}. However, the tools for interpreting MMFMs cover narrow fields. \citet{yu2024understanding, stan2024lvlm} mainly focuses on the attention mechanism in generative VLMs. \citet{aflalo2022vl} introduces a tool to visualize attentions and also hidden states of generative VLMs. \citet{joseph2023vit} proposes a tool for vision transformers, mainly focusing on attention maps, activation patches, and logit lenses. Besides, for diffusion models, \citet{João2022diffusers} provides a visualization of the inner diffusion steps of generating an image.

A unified benchmark for interpretability is also a very important research direction. In LLMs,
\citet{huang2024ravel} introduces a benchmark for evaluating interpretability methods for disentangling LLMs' representations. \citet{thurnherr2024tracrbench} presents a novel approach for generating interpretability test beds using LLMs which saves time for manually designing experimental test data. \citet{nauta2023anecdotal, schwettmann2024find} also provides benchmarks for interpretability in LLMs. However, there is no such benchmark for multimodal models, which is an important future research direction.

Overall, compared to the comprehensive tools and benchmarks in the LLMs field, there are less for multimodal foundation models. Providing a comprehensive, unified evaluation benchmark and tools is a future research direction.

\section{Main Open Challenges}
% \zihao{@Samya and @Mo, do we need this section? We deleted this section in our short version, but I keep it here right now.}
While mechanistic interpretability is a well-established and popular research area for language models, it remains in its early stages for multimodal models. This section summarizes key open challenges in the field, with a focus on downstream applications that leverage mechanistic insights. These challenges include interpreting the internal layers of diffusion transformers for tasks like model editing, extending mechanistic insights for tasks beyond VQA or simple image generation, developing sequential batch model editing techniques for multimodal models—including diffusion and multimodal language models, exploring the effectiveness of sparse autoencoders and their variants for controlling and steering multimodal models, designing transparent data attribution methods informed by mechanistic insights, and improving multimodal in-context learning through a deeper mechanistic understanding. In addition, extending mechanistic interpretability techniques to analyze unified vision-text understanding and generation models such as~\citep{xie2024show} is an open direction of research.
\section{Conclusion}

Our survey reviews mechanistic understanding methods for MMFMs, including contrastive and generative VLMs and text-to-image diffusion models, with a focus on downstream applications. We introduce a novel taxonomy differentiating interpretability methods adapted from language models and those designed for multimodal models. Additionally, we compare mechanistic insights from language and multimodal models, identifying gaps in understanding and their impact on downstream applications.

\section{Limitations}
Our work has several limitations: (1) we mainly focus on the image-text multimodal model without considering other modalities such as video, time series, or 3D. (2) We don't contain the experimental analysis because of the lack of unified benchmarks. We will consider this in our future work. (3) We only focus on the transformer-based model or diffusion model, without considering novel model architecture such as MAMBA \cite{gu2023mamba}.
\bibliography{custom}
\bibliographystyle{acl_natbib}

\newpage
\appendix

\section{More Definitions}
\label{app:more-definitions}
\paragraph{Supervision} We define a type of interpretability method as ``supervised'' if we need to have a labeled dataset to analyze it, otherwise, it is ``unsupervised''.

\begin{figure*}[ht!]
    \centering
    \includegraphics[width=0.6\textwidth]{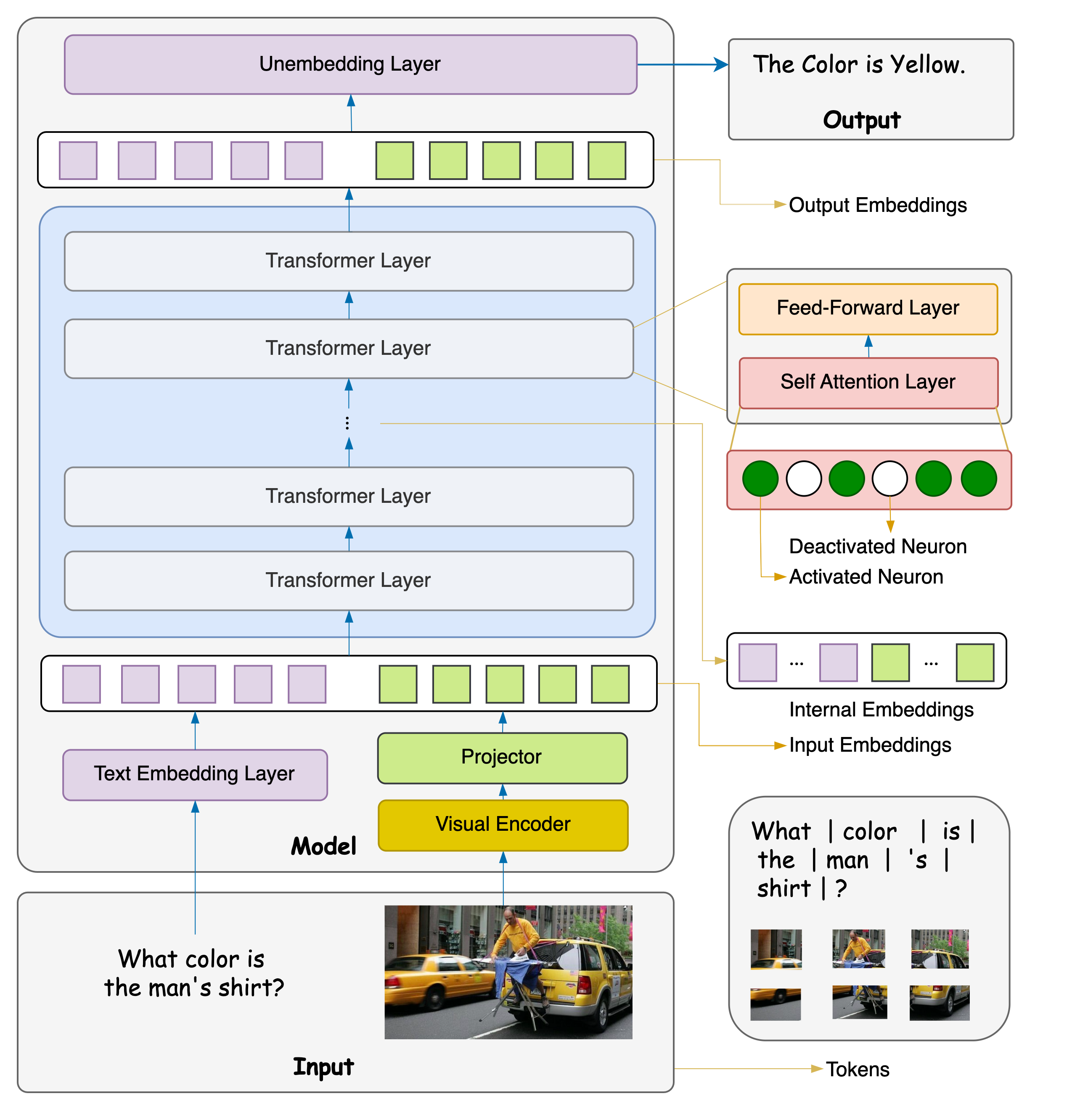}
    \caption{The illustration of model components. Take the transformer-based generative vision-language model as an example.}
    \vspace{-0.4cm}
    \label{fig:components}
\end{figure*}

In the following sections, we also classify the papers in each type of method from the following perspective: (1) the interpretability aspect - what the method aims to interpret, e.g., data influence, fine-tuning, information flow, knowledge localization, and component contribution. (2) The analyzed component of a model, e.g., emebddings, layers (MLP, self attention, cross attention), or more fine-grained neurons. The illustration of model components is shown in Figure \ref{fig:components}. (3) Applications: the downstream applications that are inspired by the insights of this method. Note, this is different from the task column in Table \ref{tab:interpretability-methods-llm-based} and \ref{tab:interpretability-method-multimodal-specific} which represents the task each paper they use to conduct interpretability analysis.

\section{More Details on LLM Interpretability Methods for Multimodal Models}

In this section, we summarize the methods from different views - Interpretability Aspect and Analyzed Component. We list the papers of Linear Probing, Logit Lens, Causal Tracing, General Task Vectors, and Sparse-Autoencoders in Table \ref{tab:linear-probing}, \ref{tab:logit-lens}, \ref{tab:causal-tracing}, \ref{tab:task-vector} and \ref{tab:sparse-autoencoder}, respectively.

In Table \ref{tab:interpretability-methods-llm-based}, we provide an overall comprehensive listing and analysis of all the papers discussed in this section. This table includes more detailed information on the datasets utilized, the models employed, and the specific tasks they conduct analysis experiments on. Note, that the ``task'' is different from ``application'' in the tables of each method, which is inspired by interpretability findings.

\begin{table*}[ht]
  \centering
  \resizebox{\textwidth}{!}{
  \begin{tabular}{clcc}
    \toprule
    Paper & Interpretability Aspect & Analyzed Component & Application \\
    \midrule
    \cite{tao2024probingmultimodallargelanguage}& Information flow& Layers & Visual-language entailment\\
    \cite{torroba-hennigen-etal-2020-intrinsic}& Knowledge localization & Neurons & Linguistic understanding\\
    \cite{dahlgren-lindstrom-etal-2020-probing}& Knowledge localization& Image-text embedding & Image-caption alignment\\
    \cite{dai2023plausiblefaithfulprobingobject}& Component contribution& Image encoding & Object hallucination\\
    \cite{cao2020scenerevealingsecretspretrained}& Information flow& Cross modal interaction & V+L benchmark\\
    \cite{salin2022vision}& Component contribution& Layers & Multimodal understanding\\
    \cite{qi2023limitation}& Data influence& Prompt & Prompt optimization\\
    \bottomrule
  \end{tabular}
  }
  \vspace{-0.2cm}
  \caption{Additional Details on Linear Probing Papers}
  \label{tab:linear-probing}
\end{table*}

\begin{table*}[ht]
  \centering
  \resizebox{\textwidth}{!}{
  \begin{tabular}{clcc}
    \toprule
    Paper & Interpretability Aspect & Analysed Component & Application \\
    \midrule
    \cite{phukan2024logitlenscontextualembeddings} & Data Influence & Hidden states & Improving VQA Performance \\
    \cite{jiang2024devilsmiddlelayerslarge} & Information flow & Attention heads & Object hallucination \\
    \cite{huo2024mmneuron} & Knowledge localization & Neurons & - \\
    \cite{zhao2024knowtokendistributionsreveal} & Information flow & Hidden states & Controllable generation \\
    \bottomrule
  \end{tabular}
  }
  \vspace{-0.2cm}
  \caption{Additional Details on Logit Len Papers}
  \label{tab:logit-lens}
\end{table*}

\begin{table*}[ht]
  \centering
  \resizebox{\textwidth}{!}{
  \begin{tabular}{clcc}
    \toprule
    Paper & Interpretability Aspect & Analysed Component & Application \\
    \midrule
    \cite{basu2023localizing}& Knowledge Localization & Self-attention & Model Editing \\
    \cite{basu2024localizing}& Knowledge Localization & Cross-attention & Model Editing \\
    \cite{basu2024understandinginformationstoragetransfer}& Knowledge Localization, Flow & MLP & Model Editing \\
    \cite{yu2024understandingmultimodalllmsmechanistic}& Knowledge Localization & Self-attention & - \\

    \cite{palit2023visionlanguagemechanisticinterpretabilitycausal}& Knowledge Localization & Self-attention & - \\
    \bottomrule
  \end{tabular}
  }
  \vspace{-0.2cm}
  \caption{Additional Details on Causal Tracing Papers}
  \label{tab:causal-tracing}
\end{table*}

\begin{table*}[ht]
  \centering
  \resizebox{\textwidth}{!}{
  \begin{tabular}{clcc}
    \toprule
    Paper & Interpretability Aspect & Analyzed Component & Application \\
    \midrule
    \cite{baumann2024continuoussubjectspecificattributecontrol}& Fine-tuning & Layers & Continuous Image Editing\\
    \cite{gandikota2025concept}& Fine-tuning & LoRA Layers & Continuous Image Editing\\
    \cite{cohenmulti}& Knowledge Localization & Layers & Model Editing\\
    \bottomrule
  \end{tabular}
  }
  \vspace{-0.2cm}
  \caption{Additional Details on General Task Vectors Papers}
  \label{tab:task-vector}
\end{table*}

\begin{table*}[ht]
  \centering
  \begin{tabular}{clcc}
    \toprule
    Paper & Interpretability Aspect & Analyzed Component & Application \\
    \midrule
    \cite{Daujotas2024}& Knowledge Localization & Layers,Neurons & Model Steering\\
    \cite{rao2024discover}& Knowledge Localization & Layers,Neurons & Model Steering\\
    \cite{lim2024sparseautoencodersrevealselective}& Knowledge Localization & Layers,Neurons & Model Steering\\

    \cite{surkov2024unpackingsdxlturbointerpreting}& Knowledge Localization & Layers,Neurons & Model Steering\\
    \cite{Sadov2024}& Knowledge Localization & Layers, Neurons & Model Steering\\
    \bottomrule
  \end{tabular}
  \vspace{-0.2cm}
  \caption{Additional Details on Sparse-Autoencoders}
  \label{tab:sparse-autoencoder}
\end{table*}

\section{More Details on Interpretability Methods Specific to Multimodal Models}
We list the papers of Text-Explanations of Internal Embeddings, Network Dissect, and Cross-Attention Interpretability in Table \ref{tab:text-explanation-internal-embedding}, \ref{tab:network-dissect} and \ref{tab:cross-attention-interpretability} respectively.

In Table \ref{tab:interpretability-method-multimodal-specific}, we provide a comprehensive listing and analysis of all the papers related to Interpretability Methods Specific to Multimodal Models.

\newpage

\begin{table*}[ht]
  \centering
  \resizebox{\textwidth}{!}{
  \begin{tabular}{clcc}
    \toprule
    Paper & Interpretability Aspect & Analyzed Component & Application \\
    \midrule
    \cite{gandelsman2024interpretingclipsimagerepresentation}& Knowledge Localization & Self-attention & Spurious Corr, Segmentation\\
    \cite{balasubramanian2024decomposing}& Knowledge Localization & Self-attention & Spurious Corr, Segmentation\\
    \cite{bhalla2024interpreting}& Knowledge Localization & Layers & Spurious Corr, Model Editing\\
    \cite{parekh2024concept}& Knowledge Localization & Self-attention & -\\

    \bottomrule
  \end{tabular}
  }
  \caption{Additional Details on Text-Explanations of Internal Embeddings Papers}
  \label{tab:text-explanation-internal-embedding}
\end{table*}

\begin{table*}[ht]
  \centering
  \resizebox{\textwidth}{!}{
  \begin{tabular}{clcc}
    \toprule
    Paper & Interpretability Aspect & Analyzed Component & Application \\
    \midrule
    \cite{falcon}& Knowledge Localization & Neurons & -\\
    \cite{oikarinen2023clipdissect}& Knowledge Localization & Embeddings & Spurious Correlation\\
    \cite{hernandez2021natural}& Knowledge Localization & Neurons & Improving Robustness for IC\\
    \cite{bai2024describe}& Knowledge Localization & Neurons & Improving Generalization for IC\\

    \bottomrule
  \end{tabular}
  }
  \caption{Additional Details on Network Dissect Papers. IC represents image classification.}
  \label{tab:network-dissect}
\end{table*}

\begin{table*}[ht]
  \centering
  \resizebox{\textwidth}{!}{
  \begin{tabular}{clcc}
    \toprule
    Paper & Interpretability Aspect & Analyzed Component & Application \\
    \midrule
    \cite{basu2024mechanistic}& Knowledge Localization & Cross-attention & Model Editing\\
    \cite{neo2024interpretingvisualinformationprocessing}& Knowledge Flow & Cross-attention & Model Editing\\
    \cite{hertz2022prompt}& Knowledge Flow & Cross-attention & Image Editing\\
    \cite{tang2022daam}& Knowledge Flow & Cross-attention & Visualization, Compositionality\\

    \bottomrule
  \end{tabular}
  }
  \caption{Additional Details on Cross-Attention Interpretability Papers}
  \label{tab:cross-attention-interpretability}
\end{table*}

\begin{table*}[t]
\centering
\scriptsize
\renewcommand{\arraystretch}{1.1}

\setlength{\tabcolsep}{3pt}

\begin{tabular}{l|p{3cm}|p{4cm}|p{2.8cm}|p{3cm}}
\hline
\textbf{Methods} & \textbf{Paper} & \textbf{Models} & \textbf{Task} & \textbf{Datasets} \\
\hline
\multirow{4}{*}{Logit Lens}
& \cite{huo2024mmneuron} & LLaVa-next, InstructBLIP & VQA & LingoQA, RS-VQA, PMC-VQA, DocVQA, VQAv2 \\
& \cite{jiang2024devilsmiddlelayerslarge} & LLaVA-1.5-7B, Shikra, MiniGPT-4 & Hallucination Detection & COCO 2014 \\

& \cite{phukan2024logitlenscontextualembeddings} & Qwen2-VL-7B, InternLM-xcomposer2-vl-7b & Hallucination Detection, VQA & High-Quality Hallucination Benchmark, TextVQA-X \\

& \cite{zhao2024knowtokendistributionsreveal} & LLaVA-v1.5 (13B/7B), InstructBLIP, mPLUG-owl & Identifying Unanswerable Questions & VizWiz, MM-SafetyBench \\
\hline
\multirow{6}{*}{Linear Probing}
& \cite{cao2020scenerevealingsecretspretrained} & ViLBERT, LXMERT, UNITER & Multimodal Fusion, Cross-modal Interaction & Visual Genome, Flickr30k \\

& \cite{dai2023plausiblefaithfulprobingobject} & OSCAR, VinVL, BLIP, OFA & Object Hallucination Detection & COCO Caption, NoCaps \\

& \cite{salin2022vision} & UNITER, LXMERT, ViLT & POS Tagging, Object Counting & Flickr30K, MS-COCO \\

& \cite{tao2024probingmultimodallargelanguage} & Kosmos-2, LaVIT, EmU, Qwen-VL & Visual-language Entailment & MS-COCO \\

& \cite{hendricks2021probingimagelanguagetransformersverb} & MMT, SMT & Verb Understanding & Conceptual Captions \\

& \cite{dahlgren-lindstrom-etal-2020-probing} & VSE++, VSE-C, HAL & Linguistic Properties & MS-COCO \\
\hline
\multirow{2}{*}{Sparse AutoEncoder}
& \cite{lim2024sparseautoencodersrevealselective} & CLIP & Image Classification & ImageNet \\

& \cite{rao2024discover} & CLIP, ResNet-50 & Concept Discovery & CC3M \\
\hline
\multirow{5}{*}{Causal Tracing}
& \cite{basu2024localizing} & Stable Diffusion, IMAGEN & Knowledge Localization & -- \\

& \cite{basu2024understandinginformationstoragetransfer} & LLaVa & VQA, Model Editing & VQA-Constraints \\

& \cite{basu2024mechanistic}& SD-XL, DeepFloyd & Knowledge Localization & -- \\

& \cite{yu2024understandingmultimodalllmsmechanistic} & LLaVa & VQA, Hallucination Detection & COCO \\

& \cite{palit2023visionlanguagemechanisticinterpretabilitycausal} & BLIP & Causal Tracing & COCO-QA \\
\hline
\multirow{3}{*}{Task Vector}
& \cite{cohenmulti} & Diffusion Model, CLIP & Multi-concept Editing & -- \\

& \cite{gandikota2025concept} & Stable Diffusion & Image Editing & Ostris Dataset, FFHQ \\

& \cite{baumann2024continuoussubjectspecificattributecontrol} & CLIP, T2I Diffusion & Image Editing & Contrastive Prompts \\
\hline
\multirow{6}{*}{In-Context Learning}
& \cite{huang2024multimodal} & Qwen-VL, Idefics2-8B & Many-shot Learning & VizWiz, OK-VQA \\

& \cite{zhou2024visual} & LLaVA, MiniGPT, Qwen-VL & Image-Content Reasoning & Emoset, CIFAR10 \\

& \cite{qin2024factors} & OpenFlamingo, GPT4V & VQA, Classification & -- \\
& \cite{mitra2025sparseattentionvectorsgenerative} & LLaVA, Qwen-VL & Classification, VQA & BLINK, NaturalBench \\

& \cite{luo2024task} & LLaVA, Mantis-Fuyu & Instruction Transfer & -- \\

& \cite{baldassini2024makes} & IDEFICS, OpenFlamingo & VQA, Captioning & COCO, VQAv2 \\
\hline
\multirow{7}{*}{Neuron-Level Description}
& \cite{huo2024mmneuron} & LLaVA-NeXT, InstructBLIP & VQA & LingoQA, RS-VQA \\

& \cite{gandelsman2024interpretingsecondordereffectsneurons} & CLIP & Zero-shot Segmentation & -- \\

& \cite{yu2024understandingmultimodalllmsmechanistic} & LLaVa & VQA & COCO \\

& \cite{tang2024language} & LLaMA-2, BLOOM & -- & -- \\

& \cite{hintersdorf2024finding} & Stable Diffusion, DALL-E & Neuron Localization & -- \\

& \cite{huang2024miner} & Qwen-VL, Qwen-Audio & -- & -- \\

& \cite{schwettmann2023multimodal} & GPT-J with BEIT & Image Captioning & CC3M \\
\hline
\end{tabular}
%} % end \scalebox if used
\caption{A comprehensive overview of interpretability methods for Section \ref{llm_interpret}.}
\label{tab:interpretability-methods-llm-based}
\end{table*}

\begin{table*}[t]
\centering
\scriptsize
\renewcommand{\arraystretch}{1.1}

\setlength{\tabcolsep}{3pt}

\begin{tabular}{l|p{3cm}|p{3cm}|p{2.8cm}|p{4cm}}
\hline
\textbf{Methods} & \textbf{Paper} & \textbf{Models} & \textbf{Task} & \textbf{Datasets} \\
\hline
\multirow{4}{*}{\makecell{Text-Explanations of\\ Internal Embeddings}}
& \cite{gandelsman2024interpretingclipsimagerepresentation}
      & CLIP
      & Image Retrieval, Segmentation
      & Waterbirds, CUB, Places, ImageNet-segmentation \\
& \cite{balasubramanian2024decomposing}
      & CLIP
      & Image Retrieval, Segmentation
      & ImageNet \\
& \cite{bhalla2024interpreting}
      & CLIP
      & Image Classification
      & CIFAR100, MIT States, MSCOCO, LAION, CelebA, ImageNetVal\\
      & \cite{parekh2024concept}
      & DePALM (CLIP+OPT)
      & Image Classification
      & COCO \\
\hline
\multirow{4}{*}{Network Dissection}
& \cite{oikarinen2023clipdissect}      & ResNet
      & Image Classification
      & CIFAR100, Broden, ImageNet \\
& \cite{falcon}
      & DINO
      & Image Classification
      & ImageNet, STL-10 \\
& \cite{hernandez2021natural}
      & ResNet, Gan, AlexNet
      & Image Classification
      & ImageNet \\
& \cite{bai2024describe}
      & ResNet
      & Image Classification
      & ImageNet \\
\hline
\multirow{18}{*}{\makecell{Training Data \\Attribution Method}}
& \cite{hu2024dissectingmisalignmentmultimodallarge@data-attribution-gm1}
      & CLIP(ViT-B/16 + LoRA)
      & ---
      & FGVC-Aircraft, Food101, Flowers102, Describable Textures Dataset(DTD), Cifar-10
      \\
      & \cite{mlodozeniec2024influencefunctionsscalabledata@data-attribution-gm3}
      & DDPM
      & ---
      & CIFAR-10, CIFAR-2, ArtBench
      \\
      & \cite{DBLP:conf/icml/ParkGILM23@data-attribution-gm4}
      & ResNet-9; ResNet-18; BERT
      & ---
      & QNLI, CIFAR-10, ImageNet
      \\
      & \cite{DBLP:conf/iclr/ZhengPD0L24@data-attribution-gm5}
      & DDPM
      & ---
      & CIFAR(32×32), CelebA(64×64), ArtBench
      \\
      & \cite{DBLP:journals/tmlr/XieLBH24@data-attribution-dm1}
      & DDPM/DDIM
      & ---
      & CIFAR-10 airplane subclass, MNIST zero subclass,
        ImageNet, CelebA, Artbench-2
      \\
      & \cite{jha2024clap4clipcontinuallearningprobabilistic@data-attribution-dm2}
      & CLIP
      & ---
      & CIFAR100, ImageNet100, ImageNet-R, CUB200, VTAB
      \\
      & \cite{DBLP:conf/nips/PruthiLKS20@data-attribution-dm3}
      & ResNet-56
      & ---
      & CIFAR-10, MNIST
      \\
      & \cite{DBLP:conf/www/QiuYCZNH022@data-attribution-pm1}
      & ResNet50, VGG16
      & ---
      & ImageNet, Pascal VOC
      \\
      & \cite{yang2024enhancingcrossprompttransferabilityvisionlanguage@data-attribution-pm2}
      & BLIP2(blip2-opt-2.7b), instructBLIP(instructblip-vicuna-7b), LLaVA(LLaVA-v1.5-7b)
      & ---
      & visualQA, CroPA
      \\
      & \cite{DBLP:conf/mm/Zheng0DL24@data-attribution-pm3}

      & CLIP
      & ---
      & Flickr30, MS COCO
      \\
      & \cite{chen2024attributionanalysismeetsmodel@data-attribution-pm4}
      & BLIP2-OPT(2.7B), LLaVA-V1.5(7B), MiniGPT-4(7B)
      & ---
      & E-VQA, E-IC
      \\
      & \cite{DBLP:conf/cvpr/MitraHDH24@data-attribution-pm5}
      & InstructBLIP-13B, LLaVA-1.5-13, Sphinx, GPT-4V
      & ---
      & Winoground, WHOOPS!, SEEDBench, MMBench, LLaVA-Bench
      \\
      & \cite{fu2024tldrtokenleveldetectivereward@data-attribution-pm6}
      & PaliGemma-3B-Mix-448
      & ---
      & DOCCI
      \\
      & \cite{DBLP:conf/iclr/KwonWW024@data-attribution-pm7}
      & RoBERTa / Llama-2-13B-chat, stable-diffusion-v1.5
      & ---
      & \makebox[0pt][l]{\parbox{6cm}{MRPC, SST2, WNLI, QQP, \\ Dreambooth (various transformations)}}
      \\
      & \cite{DBLP:conf/iccv/WangEZ0234@data-attribution-clm1}
      & DINO, MoCov3, CLIP, ViT, ALADIN, SSCD
      & ---
      & ImageNet-1K, BAM-FG, Artchive, MSCOCO
      \\
      & \cite{DBLP:conf/aaai/PengZYZQ24@data-attribution-clm2}
      & CLIP
      & ---
      & CIFAR10, CIFAR100, FGVC Aircraft, Oxfordpet, Stanford Cars, DTD, Food101, SUN397
      \\
      & \cite{DBLP:conf/aaai/PengZYZQ24@data-attribution-clm2}
      & CLIP, OpenCLIP-G/14, EVA-02-CLIP-bigE-14-plus, ALBEF, TCL, BLIP, BLIP2, MiniGPT-4
      & ---
      & MSCOCO, Flickr30K, SNLI-VE
      \\
      & \cite{DBLP:conf/nips/WangRW23@data-attribution-ibm1}
      & CLIP
      & ---
      & Conceptual Captions, MS-CXR, ROCO, RSICD
      \\
      & \cite{DBLP:conf/wacv/FangWZHZXW024@data-attribution-ibm2}
      & DensetNet-121
      & ---
      & ITAC, iCTCF, BRCA, ROSMAP
      \\
\hline
\multirow{4}{*}{\makecell{Cross-attention \\Interpretability \\Methods}}

    &\cite{basu2024mechanistic}& SD-1.5, SD-XL, DeepFloyd & Model Editing & Concept-Editing Dataset \\
    &\cite{neo2024interpretingvisualinformationprocessing}& LLaVA, LLaVA-Phi & Potential Application: Coarse Segmentation & COCO Detection Dataset\\
    &\cite{hertz2022prompt}& Stable-Diffusion & Image Editing & Custom Image Editing Dataset\\
    &\cite{tang2022daam}& Stable-Diffusion & Visualization & Custom Dataset\\
\hline
\end{tabular}
%} % end \scalebox if used
\caption{A comprehensive overview of interpretability methods for Section \ref{non_llm_interpret}.}
\label{tab:interpretability-method-multimodal-specific}
\end{table*}

\end{document}